\documentclass[lettersize,journal]{IEEEtran}
\usepackage{amsmath,amsfonts}
\usepackage{utfsym}
\usepackage{algorithmic}
\usepackage{algorithm}
\usepackage{array}
\usepackage[caption=false,font=normalsize,labelfont=sf,textfont=sf]{subfig}
\usepackage{textcomp}
\usepackage{stfloats}
\usepackage{url}
\usepackage{verbatim}
\usepackage{graphicx}
\usepackage{cite}
\usepackage{dirtytalk}
\usepackage{booktabs}
\usepackage{multirow}
\usepackage{rotating}
\usepackage{hyperref}
\usepackage{hyperref}[colorlinks,linkcolor=blue]
\usepackage{multirow}
\usepackage{makecell}
\usepackage{url}
\hypersetup{
	colorlinks=true,
	linkcolor=cyan,
	filecolor=blue,      
	urlcolor=red,
	citecolor=green,
}
\usepackage[T1]{fontenc}

\newcommand{\etal}{\textit{et al.}}

\newcommand{\etc}{\textit{etc.}}

\begin{document}

\title{A Survey of Deep Learning in Sports Applications: Perception, Comprehension, and Decision}

\author{
Zhonghan Zhao$^*$, 
Wenhao Chai$^*$, 
Shengyu Hao, 
Wenhao Hu, 
Guanhong Wang, 
Shidong Cao, 
\\Mingli Song,~\IEEEmembership{Senior Member,~IEEE,}
Jenq-Neng Hwang,~\IEEEmembership{Fellow,~IEEE,}
Gaoang Wang$^\dagger$,~\IEEEmembership{Member,~IEEE}
\thanks{$^*$ Equal contribution.}
\thanks{$^\dagger$ Corresponding author: Gaoang Wang.}
\thanks{Zhonghan Zhao, Shengyu Hao, Wenhao Hu, Guanhong Wang, Mingli Song are with College of Computer Science and Technology, Zhejiang University.}
\thanks{Shidong Cao is with the Zhejiang University-University of Illinois Urbana-Champaign Institute, Zhejiang University.}
\thanks{Gaoang Wang is with the Zhejiang University-University of Illinois Urbana-Champaign Institute, and College of Computer Science and Technology, Zhejiang University.}
\thanks{Wenhao Chai and Jenq-Neng Hwang are with the University of Washington.}
}




\maketitle

\begin{abstract}
Deep learning has the potential to revolutionize sports performance, with applications ranging from perception and comprehension to decision. This paper presents a comprehensive survey of deep learning in sports performance, focusing on three main aspects: algorithms, datasets and virtual environments, and challenges.  Firstly, we discuss the hierarchical structure of deep learning algorithms in sports performance which includes perception, comprehension and decision while comparing their strengths and weaknesses. Secondly, we list widely used existing datasets in sports and highlight their characteristics and limitations. Finally, we summarize current challenges and point out future trends of deep learning in sports. Our survey provides valuable reference material for researchers interested in deep learning in sports applications.
\end{abstract}


\section{Introduction}
\IEEEPARstart{A}{rtificial} Intelligence (AI) has found wide-ranging applications and holds a bright future in the world of sports. Its ever-growing involvement is set to revolutionize the industry in myriad ways, enabling new heights of efficiency and precision.

A prominent application of AI in sports is the use of deep learning techniques. Specifically, these advanced algorithms are utilized in areas like player performance analysis, injury prediction, and game strategy formulation~\cite{chmait2021artificial}. Through capturing and processing large amounts of data, deep learning models can predict outcomes, uncover patterns, and formulate strategies that might not be evident to the human eye. This seamless integration of deep learning and the sports industry~\cite{SMTSport38:online,vizrt} exemplifies how technology is enhancing our ability to optimize sporting performance and decision-making.

\begin{figure}[t]
    \centering
    \includegraphics[width=\linewidth]{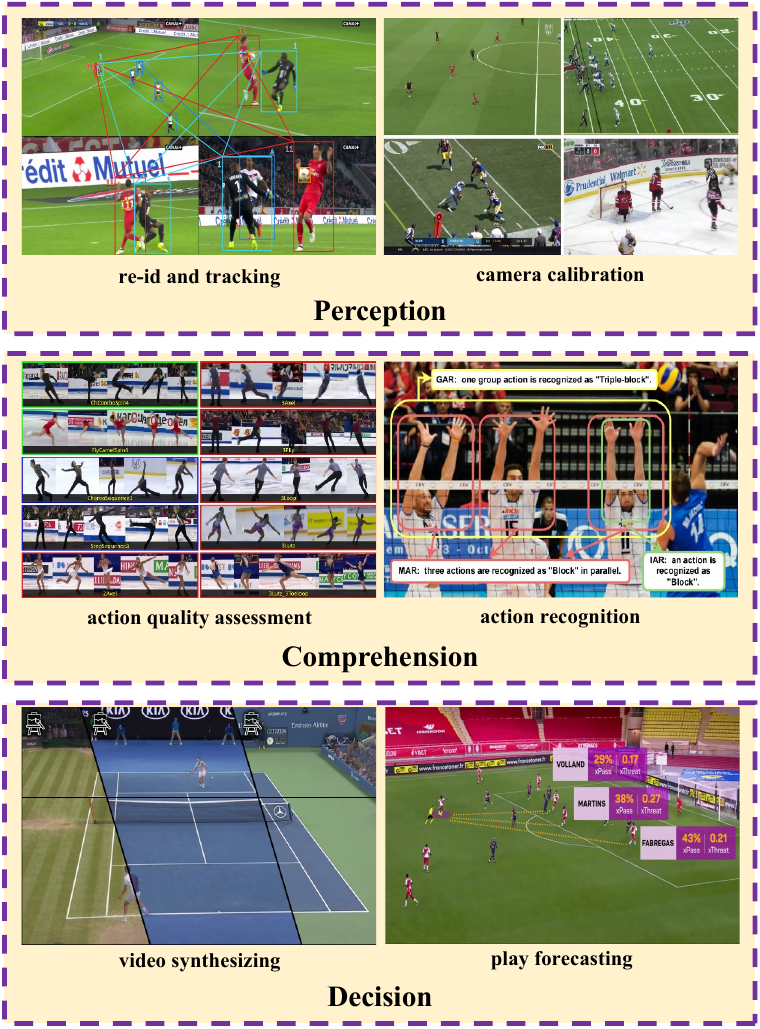}
    \caption{The examples of the applications in sports performance in perception, comprehension, and decision.}
    \label{fig:illustration}
\end{figure}

Although predicting and optimizing athletic performance has numerous advantages, it remains a complex problem. Traditionally, sports experts like coaches, managers, scouts, and sports health professionals have relied on conventional analytical methods to tackle these challenges. However, gathering statistical data and analyzing decisions manually is a demanding and time-consuming endeavor~\cite{AISPA}. Consequently, an automated system powered by machine learning emerges as a promising solution that can revolutionize the sports industry by automating the processing of large-scale data.

In recent years, there has been a notable increase in comprehensive surveys exploring the applications of machine learning and deep learning in sports performance. These surveys cover a wide range of topics, including the recognition of sports-specific movements~\cite{cust2019machine}, mining sports data~\cite{bonidia2018computational}, and employing AI techniques in team sports~\cite{beal2019artificial}. While some surveys focus on specific sports like soccer~\cite{beal2019artificial} and badminton~\cite{tan2016review}, others concentrate on particular tasks within computer vision, such as video action recognition~\cite{wu2022survey}, video action quality assessment~\cite{wang2021survey}, and ball tracking~\cite{kamble2019ball}. Furthermore, several studies explore the usage of wearable technology~\cite{adesida2019exploring, rana2020wearable} and motion capture systems~\cite{van2018accuracy} in sports, with a particular emphasis on the Internet of Things (IoT).

Previous studies~\cite{turing2009computing,wang2019ai} have employed a hierarchical approach to analyze sports performance, starting from lower-level aspects and progressing to higher-level components, while also providing training recommendations. In order to comprehend the utilization of deep learning in sports, we have segmented it into three levels: \textbf{Perception}, \textbf{Comprehension}, and \textbf{Decision}. Additionally, we have categorized diverse datasets according to specific sports disciplines and outlined the primary challenges associated with deep learning methodologies and datasets. Furthermore, we have highlighted the future directions of deep learning in motion, based on the current work built upon foundational models.

The contributions of this comprehensive survey of deep learning in sports performance can be summarized in three key aspects. 
\begin{itemize}
    \item We propose a hierarchical structure that systematically divides deep learning tasks into three categories: Perception, Comprehension, and Decision, covering low-level to high-level tasks. 
    \item We provide a summary of sports datasets and virtual environments. Meanwhile, this paper covers dozens of sports scenarios, processing both visual information and IoT sensor data. 
    \item We summarize the current challenges and future feasible research directions for deep learning in various sports fields.
\end{itemize}

The paper is organized as follows: Section \ref{sec:perception}, \ref{sec:comprehension}, and \ref{sec:decision} introduce different tasks with methods for perception, comprehension, and decision tasks in sports. Section \ref{sec:dataset} and \ref{sec:env} discuss the sports-related datasets and virtual environments. In Section \ref{sec:challenges} and \ref{sec:fw}, we highlight the current challenges and future trends of deep learning in sports. Lastly, we conclude the paper in Section \ref{sec:conc}.

\begin{figure*}[t]
    \centering
    \includegraphics[width=0.99\linewidth]{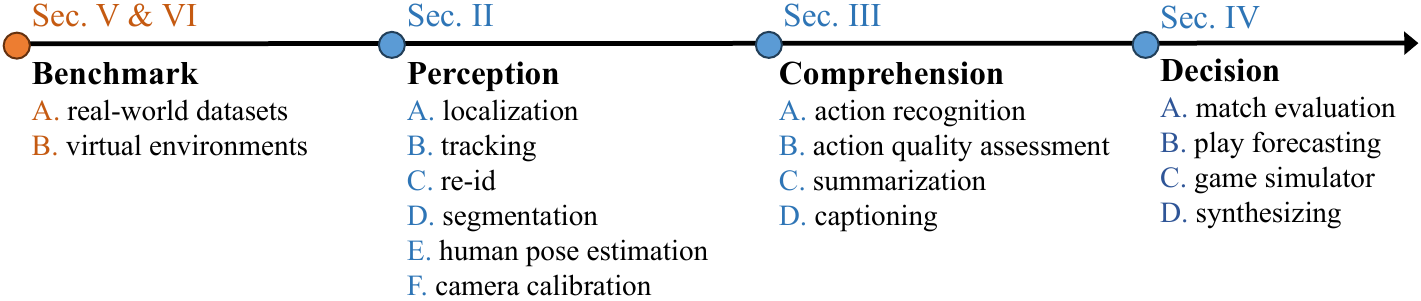}
    \caption{\textbf{Taxonomy}. A hierarchical structure that contains three categories of tasks: Perception, Comprehension, and Decision, as well as Benchmark.}
    \label{fig:taxonomy}
\end{figure*}
\section{Perception}\label{sec:perception}
\begin{figure}[!t]
    \centering
    \includegraphics[width=1\linewidth]{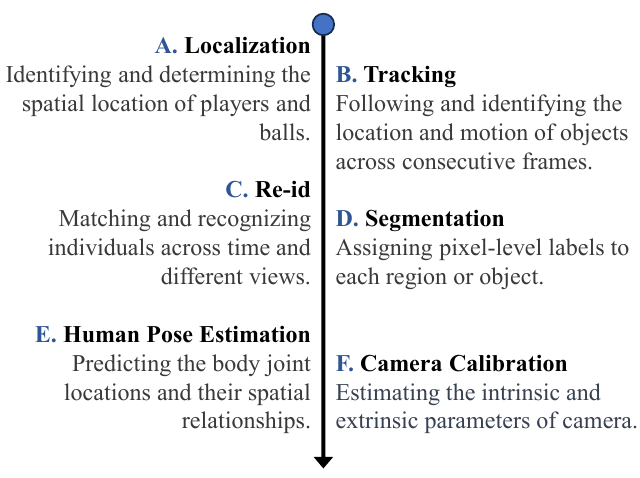}
    \caption{Taxonomy and description of perception tasks.}
    \label{fig:perception}
\end{figure}

Perception involves the fundamental interpretation of acquired data. This section presents different deep-learning methodologies tailored to specific sports tasks at the perception level as shown in Figure~\ref{fig:perception}. The subsequent perception segment will encompass tasks such as player tracking, player pose recognition, player instance segmentation, ball localization, camera calibration~\etc.

\subsection{Player and Ball Localization}
Player and ball localization aims at identifying and determining the spatial location of players and balls, which is an essential undertaking in sports video analysis. Precisely identifying these entities can provide valuable insights into team performance, enabling coaches to make well-informed decisions using data. In recent years, numerous deep learning-based techniques have emerged, specifically designed for accurately localizing players and balls in a variety of sports, such as soccer, basketball, and cricket.

\subsubsection{\bf{Player Localization}}
Player localization or detection~\cite{rao2015novel, yang20183d, csah2019evaluation} serves as a foundation for various downstream applications within the field of sports analysis. These applications include identifying player jersey numbers~\cite{gerke2017soccer, li2018jersey, liu2019pose} and teams~\cite{istasse2019associative, koshkina2021contrastive}, predicting movements and intentions~\cite{manafifard2017survey, theagarajan2018soccer, arbues2020using}. Some works~\cite{ren2015faster} leverage advancements in generic object detection to enhance the understanding of soccer broadcasts. Others~\cite{koshkina2021contrastive} focus on unsupervised methods to differentiate player teams and employ multi-modal and multi-view distillation approaches for player detection in amateur sports~\cite{cioppa2019arthus}. Vandeghen~\etal~\cite{vandeghen2022semi} introduces a distillation method for semi-supervised learning, which significantly reduces the reliance on labeled data. Moreover, certain studies~\cite{sanford2020group, cioppa2021camera} utilize player localization for action recognition and spotting. Object tracking~\cite{cioppa2022soccernet} is also crucial for the temporal localization of players.

\subsubsection{\bf{Ball Localization}}

Ball localization provides crucial 3D positional information about the ball, which offers comprehensive insights into its movement state \cite{kamble2019ball}. This task involves estimating the ball's diameter in pixels within an image patch centered on the ball, and it finds applications in various aspects of game analytics~\cite{cioppa2022scaling}. These applications include automated offside detection in soccer~\cite{uchida2021automated}, release point localization in basketball~\cite{van2022deepsportradar}, and event spotting in table tennis~\cite{voeikov2020ttnet}.

Existing solutions often rely on multi-view points~\cite{maksai2016players, cheng2017simultaneous, parisot2019consensus} to triangulate the 2D positions of the ball detected in individual frames, providing robustness against occlusions that are prevalent in team sports such as basketball or American football.

However, in single-view ball 3D localization, occlusion becomes a significant challenge. Most approaches resort to fitting 3D ballistic trajectories based on the 2D detections~\cite{skold2015estimating, parisot2019consensus}, limiting their effectiveness in detecting the ball during free fall when it follows ballistic paths. Nonetheless, in many game situations, the ball may be partially visible or fully occluded during free fall. Van~\etal~\cite{van2022deepsportradar, ball3d} address these limitations by deviating from assumptions of ballistic trajectory, time consistency, and clear visibility. They propose an image-based method that detects the ball's center and estimates its size within the image space, bridging the gap between trajectory predictions offered by ballistic approaches. Additionally, there are also works on reconstructing 3D shuttle trajectories in badminton~\cite{liu2022monotrack}.

\subsection{Player and Ball Tracking}

Player and ball tracking is the process of consistently following and identifying the location and motion of objects across consecutive frames. This tracking operation is integral to facilitating an automated understanding of sports activities. 

\subsubsection{\bf{Player Tracking}}

Tracking players in the temporal dimension is immensely valuable for gathering player-specific statistics. Recent works~\cite{naik2022deepplayer, naik2023yolov3} utilize the SORT algorithm \cite{bewley2016simple}, which combines Kalman filtering with the Hungarian algorithm to associate overlapping bounding boxes. Additionally, Hurault~\etal~\cite{hurault2020self} employ a self-supervised approach, fine-tuning an object detection model trained on generic objects specifically for soccer player detection and tracking.

In player tracking, a common challenge arises from similar appearances that make it difficult to associate detections and maintain identity consistency. Intuitively, integrating information from other tasks can assist in tracking. Some works~\cite{zhang2020multi} explore patterns in jersey numbers, team classification, and pose-guided partial features to handle player identity switches and correlate player IDs using the K-shortest path algorithm. In dance scenarios, incorporating skeleton features from human pose estimation significantly improves tracking performance in challenging scenes with uniform costumes and diverse movements~\cite{peize2021dance}.

To address identity mismatches during occlusions, Naik~\etal~\cite{naik2022deepplayer} utilize the difference in jersey color between teams and referees in soccer. They update color masks in the tracker module from frame to frame, assigning tracker IDs based on jersey color. Additionally, other works~\cite{naik2023yolov3, 8968839} tackle occlusion issues using DeepSort~\cite{wojke2017simple}.

\subsubsection{\bf{Ball Tracking}}
Accurately recognizing and tracking a high-speed, small ball from raw video poses significant challenges. Huang~\etal~\cite{8909871} propose a heatmap-based deep learning network~\cite{belagiannis2017recurrent, pfister2015flowing} to identify the ball image in a single frame and learn its flight patterns across consecutive frames. Furthermore, precise ball tracking is essential for assisting other tasks, such as recognizing spin actions in table tennis~\cite{schwarcz2019spin} by combining ball tracking information.

\subsection{Player Re-identification}\label{sec:reid}
Player re-identification~(ReID) is a task of matching and recognizing individuals across time and different views. In technical terms, this involves comparing an image of a person, referred to as the query, against a collection of other images within a large database, known as the gallery, taken from various camera viewpoints. In sports, the ReID task aims to re-identify players, coaches, and referees across images captured successively from moving cameras~\cite{van2022deepsportradar, comandur2022sports}. Challenges such as similar appearances and occlusions and the low resolution of player details in broadcast videos make player re-identification a challenging task.

Addressing these challenges, many approaches have focused on recognizing jersey numbers as a means of identifying players~\cite{liu2019pose, Nady2021PlayerII}, or have employed part-based classification techniques~\cite{Senocak2018cnnsvm}. Recently, Teket~\etal~\cite{TeketBasketball2020} proposed a real-time capable pipeline for player detection and identification using a Siamese network with a triplet loss to distinguish players from each other, without relying on fixed classes or jersey numbers. An~\etal~\cite{an2022attention} introduced a multi-granularity network with an attention mechanism for player ReID, while Habel~\etal~\cite{habel2022clip} utilized CLIP with InfoNCE loss as an objective, focusing on class-agnostic approaches.

To address the issue of low-resolution player details in multi-view soccer match broadcast videos, Comandur~\etal~\cite{comandur2022sports} proposed a model that re-identifies players by ranking replay frames based on their distance to a given action frame, incorporating a centroid loss, triplet loss, and cross-entropy loss to increase the margin between clusters.

In addition, some researchers have explored semi-supervised or weakly supervised methods. Maglo~\etal~\cite{maglo2022efficient} developed a semi-interactive system using a transformer-based architecture for player ReID. Similarly, in hockey, Vats~\etal~\cite{vats2021ice} employed a weakly-supervised training approach with cross-entropy loss to predict jersey numbers as a form of classification.

\subsection{Player Instance Segmentation}
Player instance segmentation aims at assigning pixel-level labels to each player. In player instance segmentation, occlusion is the key problem, especially in crowded regions, like basketball~\cite{van2022deepsportradar}. Some works~\cite{yan2022dual,yan2022strong} utilize online specific copy-paste method~\cite{ghiasi2021simple} to address the occlusion issue.

Moreover, instance segmentation features can be used to distinguish different players in team sports with different actions~\cite{koshkina2021contrastive, zhang2023recognition}. In hockey, Koshkina~\etal~\cite{koshkina2021contrastive} use Mask R-CNN~\cite{maskrcnn} to detect and segment each person on the playing surface. Zhang~\etal~\cite{zhang2023recognition} utilize the segmentation task to enhance throw action recognition~\cite{zhang2023recognition} and event spotting~\cite{voeikov2020ttnet}.

\subsection{Player Pose Estimation}\label{sec:est}

Player pose estimation contributes to predicting the body joint locations and their spatial relationships. It often serves as a foundational component for various tasks~\cite{chai2023global}, but there are limited works that specifically address the unique characteristics of sports scenes, such as their long processing times, reliance on appearance models, and sensitivity to calibration errors and noisy detections.

Recent approaches have employed OpenPose~\cite{cao2021openpose} for action detection or positional predictions of different elements in sports practice \cite{promrit2019model, suda2019prediction, shimizu2019prediction}. For sports with rapidly changing player movements, such as table tennis, some works~\cite{wu2020futurepong} utilize a long short-term pose prediction network~\cite{sak2014long} to ensure real-time performance. In specific actions analysis of sports videos, certain works~\cite{einfalt2019frame} use pose estimation techniques. Furthermore, Thilakarathne~\etal~\cite{thilakarathne2022pose} utilize tracked poses as input to enhance group activity recognition in volleyball. In more spatial heavy sports where less action or movement is present but more complexity lies in the poses, researchers focus on providing practitioners with tools to verify the correctness of their poses for more efficient learning, such as in Taichi~\cite{tharatipyakul2020pose} and Yoga~\cite{trejo2018recognition}.

\subsection{Camera Calibration}

Camera calibration in sports, also known as field registration, aims at estimating the intrinsic and extrinsic parameters of cameras. Homography provides a mapping between a planar field and the corresponding visible area within an image. Field calibration plays a crucial role in tasks that benefit from position information within the stadium, such as 3D player tracking on the field. Various approaches have been employed to solve sport-field registrations in different sports domains, including tennis, volleyball, and soccer~\cite{farin2003robust, yao2017fast}, often relying on keypoint retrieval methods.

With the emergence of deep learning, recent approaches focus on learning a representation of the visible sports field through various forms of semantic segmentation~\cite{homayounfar2017sports, chen2019sports, sha2020end, cioppa2021camera}. These approaches either directly predict or regress an initial homography matrix~\cite{Nie2021RobustEfficientFramework, Shi2022SelfSupervisedShape, chu2022sports}, or search for the best matching homography in a reference database~\cite{sha2020end, zhang2021high} that contains synthetic images with known homography matrices or camera parameters. In other cases~\cite{chen2019sports, sha2020end}, a dictionary of camera views is utilized, connecting an image projection of a synthetic reference field model to a homography. The segmentation is then linked to the closest synthetic view in the dictionary, providing an approximate camera parameter estimate, which is further refined for the final prediction.
\section{Comprehension}\label{sec:comprehension}

\begin{figure}[!h]
    \centering
    \includegraphics[width=1\linewidth]{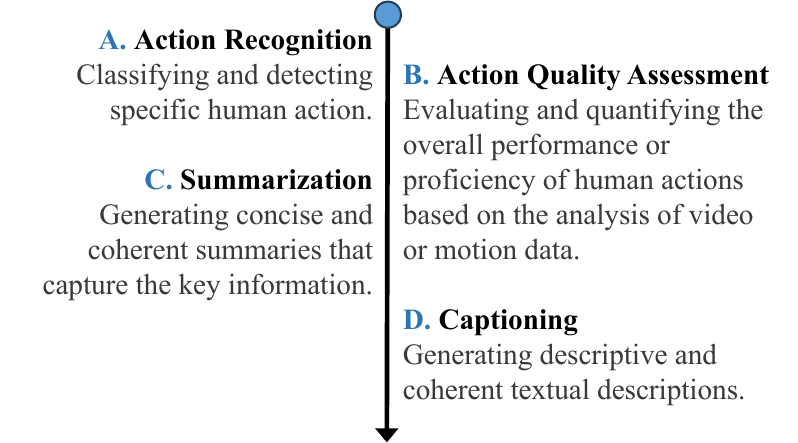}
    \caption{Taxonomy and description of comprehension tasks.}
    \label{fig:comprehension}
\end{figure}



Comprehension can be defined as the process of understanding and analyzing data. It involves higher-level tasks compared to the perception stage discussed in Section~\ref{sec:perception}. In order to achieve a comprehensive understanding of sports, the implementation can utilize raw data and directly or indirectly incorporate the tasks from the perception layer. Namely, it can utilize the outputs obtained from the perception network, such as human skeletons, depth images~\etc

In this section, we delve into specific tasks related to understanding and analyzing sports as shown in Figure~\ref{fig:comprehension}. These tasks include individual and group action recognition, action quality assessment, action spotting, sports video summarization, and captioning.


\renewcommand{\thempfootnote}{\fnsymbol{mpfootnote}}
\begin{table*}[!htp]
\caption{Deep learning models for Sports comprehension.  ``IAR'', ``GAR'', ``AQA'' stand for Individual Action Recognition, Group Action Recognition, Action Quality Assessment.}\label{tab:model}
    \centering
    \begin{tabular}{|c|c|c|l|c|}         
        \hline
        Task &Method & Venue & Benchmark & Link\\
    \hline\hline
        \multirow{13}*{IAR}
            & TSM~\cite{lin2019tsm} & ICCV-2019 &FineGym, P$^2$A &\href{https://github.com/mit-han-lab/temporal-shift-module}{$\usym{2713}$}\\ 
            \cline{2-5} & CSN~\cite{tran2019video} & ICCV-2019 &Sports 1M &\href{https://github.com/facebookresearch/VMZ}{$\usym{2713}$}\\ 
            \cline{2-5} & SlowFast~\cite{feichtenhofer2019slowfast} &ICCV-2019 &P$^2$A, Diving48 &\href{https://github.com/facebookresearch/SlowFast}{$\usym{2713}$}\\
            \cline{2-5} & G-Blend~\cite{wang2020makes} & CVPR-2020 &Sports 1M &\href{https://github.com/facebookresearch/VMZ}{$\usym{2713}$}\\
            \cline{2-5} & AGCN~\cite{shi2020skeleton} & TIP-2020 &FSD-10 &\href{https://github.com/lshiwjx/2s-AGCN}{$\usym{2713}$}\\
            \cline{2-5} & ResGCN~\cite{song2020stronger} &MM-2020 &FSD-10 &\href{https://github.com/Thomas-yx/ResGCNv1}{$\usym{2713}$}\\
            \cline{2-5} & MoViNet~\cite{kondratyuk2021movinets} & CVPR-2021 &P$^2$A &\href{https://github.com/Atze00/MoViNet-pytorch}{$\usym{2713}$}\\
            \cline{2-5} & TimeSformer~\cite{bertasius2021space} & ICML-2021 &P$^2$A, Diving48 &\href{https://github.com/facebookresearch/TimeSformer}{$\usym{2713}$}\\
            \cline{2-5} & ViSwin~\cite{liu2021video} & arXiv-2021 &P$^2$A &\href{https://github.com/SwinTransformer/Video-Swin-Transformer}{$\usym{2713}$}\\
            \cline{2-5} & ORViT~\cite{herzig2021object} & arXiv-2021 &Diving48 &\href{https://github.com/eladb3/ORViT}{$\usym{2713}$}\\
            \cline{2-5} & BEVT~\cite{wang2021bevt} & arXiv-2021 & Diving48 &\href{https://github.com/xyzforever/BEVT}{$\usym{2713}$}\\
            \cline{2-5} & VIMPAC~\cite{tan2021vimpac} & arXiv-2021 & Diving48 &\href{https://github.com/airsplay/vimpac}{$\usym{2713}$}\\ 
            \cline{2-5} & CTR-GCN~\cite{chen2021channel} &ICCV-2021 &FSD-10 &\href{https://github.com/Uason-Chen/CTR-GCN}{$\usym{2713}$}\\ 
    \hline\hline
        \multirow{2}*{GAR} 
            & DIN~\cite{yuan2021spatio} & ICCV-2021 &Diving48, HierVolleyball-v2 &\href{https://github.com/JacobYuan7/DIN-Group-Activity-Recognition-Benchmark}{$\usym{2713}$}\\
            \cline{2-5} & PoseC3D~\cite{duan2021revisiting} &CVPR-2022 &FineGym, FSD-10, HierVolleyball-v2 &\href{https://github.com/kennymckormick/pyskl}{$\usym{2713}$} \\
    \hline\hline
        \multirow{6}*{AQA}          
            &S3D~\cite{S3D} & ICIP-2018  & AQA-7 &\href{https://github.com/YeTianJHU/diving-score}{$\usym{2713}$} \\
            \cline{2-5} &C3D-LSTM~\cite{AQA-7} & WACV-2019  & AQA-7 & \href{http://rtis.oit.unlv.edu/datasets.html}{$\usym{2713}$} \\
            \cline{2-5} &C3D-AVG-MTL ~\cite{MTL-AQA} & CVPR-2019   & MTL-AQA & \href{https://github.com/ParitoshParmar/MTL-AQA}{$\usym{2713}$} \\ 
            \cline{2-5} &C3D-MSLSTM~\cite{FisV-5} & TCSVT-2020  & FisV, MIT-Skate & \href{https://github.com/loadder/MS\_LSTM.git}{$\usym{2713}$} \\
            \cline{2-5} &I3D-USDL~\cite{USDL} & CVPR-2020  & AQA-7, MTL-AQA & \href{https://github.com/nzl-thu/MUSDL}{$\usym{2713}$} \\
            \cline{2-5} &TSA~\cite{TSA-Net} & ACM MM 2021  &FR-FS, AQA-7, MTL-AQA &\href{https://github.com/Shunli-Wang/TSA-Net}{$\usym{2713}$} \\
    \hline
        
    \end{tabular}
\end{table*}

\subsection{Individual Action Recognition}

Player action recognition targets classifying and detecting specific human action. Individual action recognition is commonly used for automated statistical analysis of individual sports, such as counting the occurrences of specific actions. Moreover, it plays a crucial role in analyzing tactics, identifying key moments in matches, and tracking player activity, including metrics like running distance and performance. This analysis can assist players and coaches in identifying the essential technical factors required for achieving better results. In team sports, coaches need to monitor all players on the field and their respective actions, particularly how they execute them. Therefore, an automated system capable of tracking all these elements could greatly contribute to the players' success. However, this casts a significant challenge for computers due to the simultaneous occurrence of different actions by multiple players on the sports field, leading to issues such as occlusion and confusing scenes.

While end-to-end models~\cite{bertasius2021space, liu2021video, qi2022weakly} are commonly employed in the literature on video action recognition, they are often better suited for coarse-grained classification tasks~\cite{carreira2017quo, monfort2019moments, wang2022human}, which focus on broader categories like punches or kicks. In contrast, most sports require more fine-grained methods capable of distinguishing between specific techniques within these broader categories~\cite{shao2020finegym, sun2017taichi}.

Fine-grained action recognition within a single sport can help mitigate contextual biases present in coarse-grained tasks, making it an increasingly important research area~\cite{shao2020finegym, choi2019can, weinzaepfel2021mimetics}. Skeleton-based methods~\cite{liu2020disentangling, zhu2022fencenet, hong2021video} have gained popularity for fine-grained action recognition in body-centric sports. These approaches utilize 2D or 3D human pose as input for recognizing human actions. By representing the human skeleton as a graph with joint positions as nodes and modeling the movement as changes in these graph coordinates over time, both the spatial and temporal aspects of the action can be captured. Additionally, some works~\cite{ben2021ikea, damen2018scaling, goyal2017something} focus on fine-grained action recognition in sports that do not involve body-centric actions.

\subsection{Group Action Recognition}

Group activity recognition involves recognizing activities performed by multiple individuals or objects. It plays a significant role in automated human behavior analysis in various fields, including sports, healthcare, and surveillance. Unlike multi-player activity recognition, group / team action recognition focuses on identifying a single group action that arises from the collective actions and interactions of each player within the group. This poses greater challenges compared to individual action recognition and requires the integration of multiple computer vision techniques.

Due to the involvement of multiple players, modeling player interaction relations becomes essential in group action analysis. In general, actor interaction relations can be modeled using graph convolutional networks~(GCN) or Transformers in various methods. Transformer-based methods~\cite{gavrilyuk2020actor, hu2020progressive, yan2020social, ehsanpour2020joint, pramono2020empowering, li2021groupformer} often explicitly represent spatiotemporal relations and employ attention-based techniques to model individual relations for inferring group activity. GCN-based methods~\cite{wu2019learning, yuan2021spatio} construct relational graphs of the actors and simultaneously explore spatial and temporal actor interactions using graph convolution networks.

Among them, Yan~\etal~\cite{yan2020social} construct separate spatial and temporal relation graphs to model actor relations. Gavrilyuk~\etal~\cite{gavrilyuk2020actor} encode temporal information using I3D \cite{carreira2017quo} and establish spatial relations among actors using a vanilla transformer. Li~\etal~\cite{li2021groupformer} introduces a cluster attention mechanism. Dual-AI~\cite{han2022dual} proposes a dual-path role interaction framework for group behavior recognition, incorporating temporal encoding of the actor into the transformer architecture. Moreover, the use of simple multi-layer perceptrons~(MLP) for feature extraction in group activity analysis~\cite{xu2023mlp} is an emerging approach with great potential.

Moreover, some other works focus more on specific action recognition through temporal localization rather than classification. Several automated methods have been proposed to identify important actions in a game by analyzing camera shots or semantic information. Studies~\cite{bettadapura2016leveraging, heilbron2017scc, felsen2017will} have explored human activity localization in sports videos, salient game action identification~\cite{cioppa2018bottom, tsunoda2017football}, and automatic identification and summarization of game highlights~\cite{cai2019temporal, sanabria2019deep, turchini2019flexible}. Recent methods are more on soccer. For instance, Giancola~\etal~\cite{giancola2018soccernet} introduce the concept of accurately identifying and localizing specific actions within uncut soccer broadcast videos. More recently, innovative methodologies have emerged in this field, aiming to automate the process. Cioppa~\etal~\cite{cioppa2020context} propose the application of a context-aware loss function to enhance model performance. They later demonstrated how integrating camera calibration and player localization features can improve spotting capabilities~\cite{cioppa2021camera}. Hong~\etal~\cite{hong2022spotting} propose an efficient end-to-end training approach, while Darwish~\etal~\cite{darwish2022ste} utilize spatiotemporal encoders. Alternative strategies, such as graph-based techniques~\cite{cartas2022graph} and transformer-based methods~\cite{zhu2022transformer}, offer fresh perspectives, particularly in handling relational data and addressing long-range dependencies. Lastly, Soares~\etal~\cite{soares2022action, soares2022temporally} have highlighted the potential of anchor-based methods in precise action localization and categorization.

\subsection{Action Quality Assessment}

Action quality assessment (AQA) is a method used to evaluate and quantify the overall performance or proficiency of human actions based on the analysis of video or motion data. AQA takes into account criteria such as technique, speed, and control to assess the movement and assign a score, which can be used to guide training and rehabilitation programs. AQA has proven to be reliable and valid for assessing movement quality across various sports. Research in this field primarily focuses on analyzing the actions of athletes in the Olympic Games, such as diving, gymnastics, and other sports mentioned in Section~\ref{sec:dataset}. Existing methods typically approach AQA as a regression task using various video representations supervised by scores.

Some studies concentrate on enhancing network structures to extract more distinct features. For instance, Xu~\etal~\cite{FisV-5} propose self-attentive LSTM and multi-scale convolutional skip LSTM models to predict Total Element Score~(TES) and Total Program Component Score~(PCS) in figure skating by capturing local and global sequential information in long-term videos. Xiang~\etal~\cite{S3D} divide the diving process into four stages and employ four independent P3D models for feature extraction. Pan~\etal\cite{joint-relation-graphs} develop a graph-based joint relation model that analyzes human node motion using the joint commonality module and the joint difference module. Parisi~\etal~\cite{RO-MAN} propose a recurrent neural network with a growing self-organizing structure to learn body motion sequences and facilitate matching. Kim et al. \cite{EvaluationNet} model the action as a structured process and encode action units using an LSTM network. Wang~\etal~\cite{TSA-Net} introduce a tube self-attention module for feature aggregation, enabling efficient generation of spatial-temporal contextual information through sparse feature interactions. Yu~\etal~\cite{yu2021group} construct a contrastive regression framework based on video-level features to rank videos and predict accurate scores.

Other studies focus on improving the performance of action quality assessment by designing network loss functions. Li~\etal~\cite{e2e} propose an end-to-end framework that employs C3D as a feature extractor and integrates a ranking loss with the mean squared error~(MSE) loss. Parmar~\etal~\cite{MTL-AQA} explore the AQA model in a multi-task learning scenario by introducing three parallel prediction tasks: action recognition, comment generation, and AQA score regression. Tang~\etal~\cite{USDL} propose an uncertainty-aware score distribution learning approach that takes into account difficulty levels during the modeling process, resulting in a more realistic simulation of the scoring process.

Furthermore, some studies focus on comparing the quality of paired actions. Bertasius~\etal~\cite{BPAD} propose a model for basketball games based on first-person perspective videos, utilizing a convolutional-LSTM network to detect events and evaluate the quality of any two movements.

\subsection{{Sports Video Summarization}}



Sports video summarization aims at generating concise and coherent summaries that capture the key information. It often prioritizes the recognition of player actions~\cite{8695329}. This research field aims to generate highlights of broadcasted sports videos, as these videos are often too lengthy for audiences to watch in their entirety. Given that many sports matches can have durations of 90-180 minutes, it becomes a challenging task to create a summary that includes only the most interesting and exciting events.

Agyeman~\etal~\cite{8695329} employ a 3D ResNet CNN and LSTM-based deep model to detect five different soccer sports action classes. Rafiq~\etal~\cite{2020} propose a transfer learning-based classification framework for categorizing cricket match clips~\cite{Khan2020ContentAwareSO} into five classes, utilizing a pre-trained AlexNet CNN and data augmentation. Shingrakhia~\etal~\cite{10.1007/s00371-021-02111-8} present a multimodal hybrid approach for classifying sports video segments, utilizing the hybrid rotation forest deep belief network and a stacked RNN with deep attention for the identification of key events. Li~\etal~\cite{10.1145/3485472} propose a supervised action proposal guided Q-learning based hierarchical refinement approach for structure-adaptive summarization of soccer videos. While current research in sports video summarization focuses on specific sports, further efforts are needed to develop a generic framework that can support different types of sports videos.

\subsection{{Captioning}}
Sports video captioning involves generating descriptive and coherent textual descriptions. Sports video captioning models are designed to generate sentences that provide specific details related to a particular sport, which is a multimodal~\cite{chai2022deep} task. For instance, in basketball, Yu~\etal~\cite{yu2018fine} propose a structure that consists of a CNN model for categorizing pixels into classes such as the ball, teams, and background, a model that captures player movements using optical flow features, and a component that models player relationships. These components are combined in a hierarchical structure to generate captions for NBA basketball videos. Similarly, attention mechanisms and hierarchical recurrent neural networks have been employed for captioning volleyball videos~\cite{qi2019sports}.

Furthermore, the utilization of multiple modalities can be extended to explore the creation of detailed captions or narratives for sports videos. Qi~\etal~\cite{8733019} and Yu~\etal~\cite{8578727} have successfully generated fine-grained textual descriptions for sports videos by incorporating attention mechanisms that consider motion modeling and contextual information related to groups and relationships.
\section{Decision}\label{sec:decision}
\begin{figure}[!t]
    \centering
    \includegraphics[width=1\linewidth]{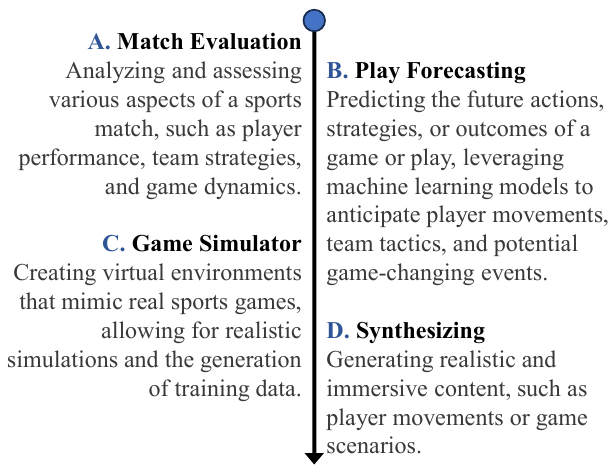}
    \caption{Taxonomy and description of decision tasks.}
    \label{fig:decision}
\end{figure}

The decision or decision-making process in sports involves the highest level of tasks, where the deployment or implicit perception and understanding of sports are essential before generating more abstract decisions. This section encompasses various tasks such as match evaluation, play forecasting, game simulation, player motion generation, and match generation as shown in Figure~\ref{fig:decision}.

\subsection{Match Evaluation}\label{sec:GA}
Match evaluation involves analyzing and assessing various aspects of a sport match, such as player performance, team strategies, and game dynamics. This task requires match modeling, often employing deep reinforcement learning methods. For instance, Wang~\etal~\cite{wang2018advantage} develop a deep reinforcement learning model to study NBA games with the goal of minimizing offensive scores. Luo~\etal~\cite{luo2020inverse} combine Q-function learning and inverse reinforcement learning to devise a unique ranking method and an alternating learning framework for a multi-agent ice hockey Markov game. Liu~\etal~\cite{liu2018deep} value player actions under different game contexts using Q-function learning and introduce a new player evaluation metric called the Game Impact Metric. Yanai~\etal~\cite{yanai2022q} model basketball games by extending the DDPG~\cite{lillicrap2015continuous} architecture to evaluate the performance of players and teams.

\subsection{Play Forecasting}
Play Forecasting aims at predicting the future actions, strategies, or outcomes of a game or play, leveraging machine learning models to anticipate player movements, team tactics, and potential game changing events. The availability of accurate player and ball tracking data in professional sports venues has generated interest in assisting coaches and analysts with data-driven predictive models of player or team behavior~\cite{stats,sec}. Several studies have utilized multiple years of match data to predict various aspects, such as predicting the ball placement in tennis~\cite{wei2016forecasting, fernando2019memory} and the likelihood of winning a point \cite{wei2016thin}. Le~\etal~\cite{le2017data} focus on predicting how NBA defenses will react to different offensive plays, while Power~\etal~\cite{power2017not} analyze the risk-reward of passes in soccer. In a more recent work, Wang~\etal~\cite{wang2021shuttlenet} delve into the analysis of where and what strokes to return in badminton.

\subsection{Game Simulators}
Game simulators typically aim at creating virtual environments that mimic real sports games, allowing for realistic simulations and the generation of training data~\cite{martins2021rsoccer,liu2019emergent,liu2021motor,kurach2020google}. These virtual environments, which are discussed in detail in Section~\ref{sec:env}, allow agents to move freely based on specific algorithms, simulating real-world sports scenarios. Within such environments, deep reinforcement learning~(DRL) algorithms have shown remarkable performance in sport-related tasks. Zhao~\etal~\cite{zhao2019multi} propose a hierarchical learning approach within a multi-agent reinforcement framework to emulate human performance in sports games. Jia~\etal~\cite{jia2020fever} address the challenges of asynchronous real-time scenarios in a basketball sports environment, supporting both single-agent and multi-agent training.

The soccer virtual environment GFootball has gained significant attention in recent years~\cite{kurach2020google}. In the 2020 Google Research Football Competition, the winning team, WeKick~\cite{wekick}, developed a powerful agent using imitation learning and distributed league training. However, WeKick is specifically designed for single-agent AI and cannot be extended to multi-agent control. To address this limitation, Huang~\etal~\cite{huang2021tikick} propose TiKick, an offline multi-agent algorithm that completes full games in GFootball using replay data generated by WeKick~\cite{huang2021tikick}. Another approach, Tizero~\cite{lin2023tizero}, trains agents from scratch without pre-collected data and employs a self-improvement process to develop high-quality AI for multi-agent control~\cite{lin2023tizero}.

Although DRL systems have made significant progress, they continue to encounter challenges in several areas, including multi-agent coordination, long-term planning, and non-transitivity~\cite{yu2021surprising,wen2022multi,ecoffet2021first}. These challenges highlight the complexity of developing AI systems that can effectively coordinate with multiple agents, make strategic decisions over extended periods, and account for non-transitive relationships in dynamic environments. Further research and advancements in these areas are crucial for enhancing the capabilities of DRL systems.

\subsection{Player Motion Synthesizing}\label{sec:PMG}

Utilizing video-based sequences to capture and analyze player movements represents a powerful approach to enhancing data diversity in sports. This innovative initiative has the potential to make a positive impact on the development of sports disciplines. Through detailed analysis and reproduction of player movements, we can gain valuable insights that have the potential to improve techniques, elevate athletic performance, and drive progress in the world of sports. This pioneering endeavor holds great promise for advancing the field and benefiting athletes and sports enthusiasts alike.

\subsubsection{\bf{Auto Choreographer}}

Creating choreography involves the creative design of dance movements. However, automating the choreography process computationally is a challenging task. It requires generating continuous and complex motion that captures the intricate relationship with accompanying music.

Music-to-dance motion generation can be approached from both 2D and 3D perspectives. 2D approaches~\cite{tendulkar2020feel,ren2020self,ferreira2021learning} rely on accurate 2D pose detectors~\cite{cao2017realtime} but have limitations in terms of expressiveness and downstream applications. On the other hand, 3D dance generation methods utilize techniques such as LSTMs~\cite{alemi2017groovenet,tang2018dance,yalta2019weakly,zhuang2020towards,kao2020temporally}, GANs~\cite{lee2019dancing,sun2020deepdance}, transformer encoders with the RNN decoder~\cite{huang2020dance} or transformer decoder~\cite{li2021learn}, and convolutional sequence-to-sequence models~\cite{ahn2020generative,ye2020choreonet} to generate motion from audio.

Early works~\cite{ren2020self,kao2020temporally,ye2020choreonet} in this field could predict future motion deterministically from audio but struggled when the same audio had multiple corresponding motions. However, recent advancements, such as the work by Li~\etal~\cite{li2021learn}, have addressed this limitation by formulating the problem with seed motion. This enables the generation of multiple motions from the same audio, even with a deterministic model. Li~\etal~\cite{li2021learn} propose a novel cross-modal transformer-based model that better preserves the correlation between music and 3D motion. This approach results in more realistic and globally translated long human motion.





\subsection{Sport Video Synthesizing}\label{sec:MG}


The goal of artificially synthesizing sports videos is to generate realistic and immersive content, such as player movements or game scenarios. Early works in this field train models using annotated videos where each time step is labeled with the corresponding action. However, these approaches use a discrete representation of actions, which make it challenging to define prior knowledge for real-world environments. Additionally, devising a suitable continuous action representation for an environment is also complex. To address the complexity of action representation in tennis, Menapace~\etal~\cite{menapace2021pvg} propose a discrete action representation. Building upon this idea, Huang~\etal~\cite{huang2022layered} model actions as a learned set of geometric transformations. Davtyan~\etal~\cite{davtyan2022glass} take a different approach by separating actions into a global shift component and a local discrete action component. More recent works in tennis have utilized a NeRF-based renderer~\cite{mildenhall2020nerf}, which allows for the representation of complex 3D scenes. Among these works, Menapace~\etal~\cite{menapace2023plotting} employ a text-based action representation that provides precise details about the specific ball-hitting action being performed and the destination of the ball.
\section{Datasets and Benchmarks}\label{sec:dataset}


In the era of deep learning, having access to effective data is crucial for training and evaluating models. In order to facilitate this, we have compiled a list of commonly used public sports datasets, along with their corresponding details, as shown in Table~\ref{tab:dataset}. Below, we provide a more detailed description of each dataset.


\begin{table*}[!htp]
    \caption{A list of video-based sports-related datasets used in the published papers. Note that some of them are not publicly available and ``multiple'' means that the dataset contains various sports instead of only one specific type of sports. ``det.'', ``cls.'', ``tra.'', ``ass.'', ``seg.'', ``loc.'',``cal.'', ``cap.'' stand for player/ball detection, action classification, player/ball tracking, action quality assessment, object segmentation, temporal action localization, camera calibration, and captioning respectively.}
    \centering
    \begin{tabular}{|c|c|c|c|c|c|c|}
    \hline
        Sport & Dataset & Year & Task & \# Videos & Avg. length & Link \\
    \hline\hline
        \multirow{8}*{Soccer} &SoccerNet~\cite{giancola2018soccernet}&2018&loc.\& cls.&500&5,400&\href{https://silviogiancola.github.io/SoccerNet/}{$\usym{2713}$}\\ 
        \cline{2-7} &SSET~\cite{feng2020sset} & 2020 &det.\&tra.  &350 &0.8h & \href{http://media.hust.edu.cn/dataset.htm}{$\usym{2713}$}\\ 
        \cline{2-7} &SoccerDB~\cite{jiang2020soccerdb} & 2020 &cls.\& loc. &346 &1.5h& \href{https://github.com/newsdata/SoccerDB}{$\usym{2713}$} \\ 
        \cline{2-7} &SoccerNet-v2~\cite{deliege2021soccernet} & 2021 &cls.\&loc.  & 500 &1.5h1.5h&\href{https://www.soccer-net.org/data}{$\usym{2713}$}\\ 
        \cline{2-7} &SoccerKicks~\cite{lessa2021soccerkicks} & 2021 &pos. &38 &- & \href{https://github.com/larocs/SoccerKicks}{$\usym{2713}$}\\ 
        \cline{2-7} &SoccerNet-v3~\cite{cioppa2022scaling} & 2022 &cls.\&tra. &346 &1.5h& \href{https://www.soccer-net.org/data}{$\usym{2713}$} \\ 
        \cline{2-7} &SoccerNet-Tracking~\cite{cioppa2022soccernet} & 2022&cls.\&tra.&21&45.5m&\href{https://www.soccer-net.org/data}{$\usym{2713}$}\\ 
        \cline{2-7} &SoccerTrack~\cite{scott2022soccertrack} & 2022 &tra.\&loc. &20&30s&\href{https://github.com/AtomScott/SoccerTrack}{$\usym{2713}$}\\ 
    \hline\hline
        \multirow{5}*{Basketball} 
        & BPAD~\cite{parisot2017scene} &2017 & ass. &48 &13m &\href{https://www.kaggle.com/datasets/gabrielvanzandycke/spiroudome-dataset}{$\usym{2713}$} \\
        \cline{2-7} & NBA~\cite{yan2020social} &2020 & cls. &181&-&\href{https://ruiyan1995.github.io/SAM.html}{$\usym{2713}$} \\  
        \cline{2-7} & NPUBasketball~\cite{ma2021npu}&2021&cls.&2,169&-&\href{https://github.com/Medjed46/NPU-RGBD-Basketball-Dataset}{$\usym{2713}$} \\ 
        \cline{2-7} & DeepSportradar-v1~\cite{van2022deepsportradar}&2022&seq.\&cal.&-&-&\href{https://github.com/DeepSportRadar}{$\usym{2713}$} \\
        \cline{2-7} & NSVA~\cite{dew2022sports}&2022&cls.\&cap.&32,019&9.5s&\href{https://github.com/jackwu502/NSVA}{$\usym{2713}$} \\
    \hline\hline
        \multirow{2}*{Tennis} 
        & PE-Tennis~\cite{menapace2022playable} & 2022 &det.\&cal. & 14,053 &3s & \href{https://github.com/willi-menapace/PlayableEnvironments}{$\usym{2713}$} \\ 
        \cline{2-7} & LGEs-Tennis~\cite{menapace2023plotting} & 2023 &cal.\&tra.\&cap. & 7,112 &7.8s& \href{https://learnable-game-engines.github.io/lge-website/}{$\usym{2713}$} \\ 
    \hline\hline
        \multirow{2}*{Figure Skating} 
        & FisV-5~\cite{xu2019learning} & 2020 & ass.\& cls. &500&2m50s&\href{https://github.com/loadder/MS_LSTM}{$\usym{2713}$} \\ 
        \cline{2-7} & FR-FS~\cite{wang2021tsa} & 2021 & ass.\& cls. & 417 & -&\href{https://github.com/Shunli-Wang/TSA-Net}{$\usym{2713}$} \\
    \hline\hline
        \multirow{2}*{Diving} & MTL-AQA~\cite{parmar2019and} &2019 & ass. &1,412&- &\href{http://rtis.oit.unlv.edu/datasets.html}{$\usym{2713}$}\\ 
        \cline{2-7} & FineDiving~\cite{xu2022finediving} &2022 & ass.\& cls. &3,000&52s &\href{https://github.com/xujinglin/FineDiving}{$\usym{2713}$}\\ 
    \hline\hline
        \multirow{5}*{Dance} & GrooveNet~\cite{alemi2017groovenet} &2017 &pos. &2 &11.5m&\href{https://omid.al/groovenet-material-ml4c/}{$\usym{2713}$} \\ 
        \cline{2-7} & Dance with Melody~\cite{tang2018dance} &2018 &pos. &61 &92s&\href{https://github.com/Music-to-dance-motion-synthesis/dataset}{$\usym{2713}$} \\
        \cline{2-7} & EA-MUD~\cite{sun2020deepdance} &2020 &pos. &17 &74s&\href{https://github.com/computer-animation-perception-group/DeepDance}{$\usym{2713}$} \\        
        \cline{2-7} & AIST++~\cite{li2021learn} &2021 &det\&pos. &1,408 &13s&\href{https://google.github.io/aistplusplus_dataset/factsfigures.html}{$\usym{2713}$} \\ 
        \cline{2-7} & DanceTrack~\cite{peize2021dance} &2022 &tra. &100 &52.9s&\href{https://github.com/DanceTrack/DanceTrack}{$\usym{2713}$} \\ 
    \hline\hline
        \multirow{1}*{Golf} & GolfDB~\cite{mcnally2019golfdb} &2019 & cls. & 1,400 &-&\href{https://github.com/wmcnally/GolfDB}{$\usym{2713}$} \\   
    \hline\hline
        \multirow{1}*{Gymnastics} & FineGym~\cite{shao2020finegym} & 2020 & cls.\& loc. & - &-&\href{https://sdolivia.github.io/FineGym/}{$\usym{2713}$}\\ 
    \hline\hline
        \multirow{1}*{Rugby} & Rugby sevens~\cite{zhu2022fencenet} &2022 & tra. &346 &40s&\href{https://kalisteo.cea.fr/index.php/free-resources/}{$\usym{2713}$} \\
    \hline\hline
        \multirow{1}*{Baseball} & MLB-YouTube\cite{piergiovanni2018fine} &2018 & cls. & 5,111 &-&\href{https://github.com/piergiaj/mlb-youtube/}{$\usym{2713}$} \\
    \hline\hline
        \multirow{14}*{General} & Sports 1M~\cite{karpathy2014large} &2014&cls.&1M&36s&\href{https://code.google.com/archive/p/sports-1m-dataset/}{$\usym{2713}$} \\ 
        \cline{2-7} & OlympicSports~\cite{pirsiavash2014assessing} &2014 &ass. &309 &-&\href{https://redirect.cs.umbc.edu/~hpirsiav/quality.html}{$\usym{2713}$} \\ 
        \cline{2-7} & SVW~\cite{safdarnejad2015sports} &2015 &det.\& cls. &4,100 &11.6s&\href{http://cvlab.cse.msu.edu/project-svw.html}{$\usym{2713}$} \\ 
        \cline{2-7} & OlympicScoring~\cite{parmar2017learning} &2017 &ass. &716 &-&\href{http://rtis.oit.unlv.edu/datasets.html}{$\usym{2713}$} \\
        \cline{2-7} & MADS~\cite{zhang2017martial} &2017 &ass. &30 &-&\href{http://visal.cs.cityu.edu.hk/research/mads/}{$\usym{2713}$} \\ 
         \cline{2-7} & MultiTHUMOS~\cite{yeung2018every} &2017&cls.&400 &4.5m&\href{http://ai.stanford.edu/~syyeung/everymoment.html}{$\usym{2713}$} \\ 
        \cline{2-7} & AQA-7~\cite{parmar2019action} &2019 &ass. &1,189 &-&\href{http://rtis.oit.unlv.edu/datasets.html}{$\usym{2713}$} \\ 
        \cline{2-7} & C-Sports~\cite{zalluhoglu2020collective} &2020 &cls.\&loc. &2,187 &-&\href{https://cemilzalluhoglu.github.io/csports}{$\usym{2713}$}\\ 
        \cline{2-7} & MultiSports~\cite{li2021multisports} &2021 &cls.\&loc.  &3,200 &20.9s&\href{https://github.com/MCG-NJU/MultiSports/}{$\usym{2713}$}\\ 
        \cline{2-7} & ASPset-510~\cite{nibali2021aspset} &2021 &pos. &510 &-&\href{https://github.com/anibali/aspset-510}{$\usym{2713}$} \\ 
        \cline{2-7} & HAA-500~\cite{chung2021haa500} &2021 &cls. &10,000 &2.12s&\href{https://www.cse.ust.hk/haa/}{$\usym{2713}$} \\ 
        \cline{2-7} & SMART~\cite{chen2021sportscap} &2021 &cls. &5,000 &-&\href{https://chenxin.tech/files/Paper/IJCV2020_Sport/project_page_SportsCap/index.htm}{$\usym{2713}$} \\ 
        \cline{2-7} & Win-Fail~\cite{parmar2022win} &2022 &cls. &817 &3.3s&\href{https://github.com/ParitoshParmar/Win-Fail-Action-Recognition}{$\usym{2713}$} \\
        \cline{2-7} & SportsPose~\cite{ingwersen2023sportspose} &2023 &pos. &25 &11m&\href{https://github.com/ChristianIngwersen/SportsPose}{$\usym{2713}$} \\
        \cline{2-7} & SportsMOT~\cite{cui2023sportsmot} &2023 &tra. &240 &25s&\href{https://deeperaction.github.io/datasets/sportsmot.html}{$\usym{2713}$} \\
    \hline
    \end{tabular}
    \label{tab:dataset}
\end{table*}

\subsection{Soccer}

In soccer, most video-based datasets benefit from active tasks like player tracking and action recognition, while some datasets focus on field localization and registration or player depth maps and meshes.

Some datasets focus more on player detection and tracking. Soccer-ISSIA~\cite{d2009semi} is an early work and a relatively small dataset with player bounding box annotations. SVPP~\cite{pettersen2014soccer} provides a multi-sensor dataset that includes body sensor data and video data. Soccer Player~\cite{lu2017light} is specifically designed for player detection and tracking, while SoccerTrack~\cite{scott2022soccertrack} is a novel dataset with multi-view and super high definition.

Other datasets like Football Action~\cite{tsunoda2017football} and SoccerDB~\cite{jiang2020soccerdb} benefit action recognition, and ComprehensiveSoccer~\cite{yu2018comprehensive} and SSET~\cite{feng2020sset} can be used for various video analysis tasks, such as action classification, localization, and player detection. SoccerKicks~\cite{deliege2021soccernet} provides player pose estimation. GOAL~\cite{qi2023goal} supports knowledge-grounded video captioning.

The SoccerNet series~\cite{giancola2018soccernet,deliege2021soccernet,cioppa2022scaling,cioppa2022soccernet} is the largest one including annotations for a variety of spatial annotations and cross-view correspondences. It covers multiple vision-based tasks including player understanding like player tracking, re-identification, broadcast video understanding like action spotting, video captioning, and field understanding like camera calibration. 

In recent years, the combination of large-scale datasets and deep learning models has become increasingly popular in the field of soccer tasks, raising the popularity of the SoccerNet series datasets~\cite{giancola2018soccernet,deliege2021soccernet,cioppa2022scaling}. Meanwhile, SoccerDB~\cite{jiang2020soccerdb}, SSET~\cite{feng2020sset}, and ComprehensiveSoccer~\cite{yu2018comprehensive} are more suitable for tasks that require player detection. However, there are few datasets like SoccerKick~\cite{lessa2021soccerkicks} for soccer player pose estimation. It is hoped that more attention can be paid to the recognition and understanding of player skeletal movements in the future.

\subsection{Basketball} 

Basketball datasets have been developed for various tasks such as player and ball detection, action recognition, and pose estimation. APIDIS~\cite{de2008distributed,parisot2019consensus} is a challenging dataset with annotations for player and ball positions, and clock and non-clock actions. Basket-1,2~\cite{maksai2016players} consists of two frame sequences for action recognition and ball detection. NCAA~\cite{ramanathan2016detecting} is a large dataset with action categories and bounding boxes for player detection. SPIROUDOME~\cite{parisot2017scene} focuses on player detection and localization. BPAD~\cite{BPAD} is a first-person perspective dataset with labeled basketball events. SpaceJam~\cite{francia2018classificazione} is for action recognition with estimated player poses. FineBasketball~\cite{gu2020fine} is a fine-grained dataset with 3 broad and 26 fine-grained categories. NBA~\cite{yan2020social} is a dataset for group activity recognition, where each clip belongs to one of the nine group activities, and no individual annotations are provided, such as separate action labels and bounding boxes. NPUBasketball~\cite{ma2021npu} contains RGB frames, depth maps, and skeleton information for various types of action recognition models. DeepSportradar-v1~\cite{van2022deepsportradar} is a multi-label dataset for 3D localization, calibration, and instance segmentation tasks.
In Captioning task, NSVA~\cite{dew2022sports} is the largest open-source dataset in the basketball domain. Compared to SVN~\cite{yan2019fine} and SVCDV~\cite{qi2019sports}, NSVA is publicly accessible and has the most sentences among the three datasets, with five times more videos than both SVN and SVCDV.
Additionally, there are some special datasets that focus on reconstructing the player. NBA2K dataset~\cite{zhu_2020_eccv_nba} includes body meshes and texture data of several NBA players.

\subsection{Volleyball}

Despite being a popular sport, there are only a few volleyball datasets available, most of which are on small scales. Volleyball-1,2~\cite{maksai2016players} contains two sequences with manually annotated ball positions. HierVolleyball~\cite{ibrahim2016hierarchical} and its extension HierVolleyball-v2~\cite{DBLP:journals/corr/IbrahimMDVM16} are developed for team activity recognition, with annotated player actions and positions. Sports Video Captioning Dataset-Volleyball~(SVCDV)~\cite{qi2019sports} is a dataset for captioning tasks, with 55 videos from YouTube, each containing an average of 9.2 sentences. However, this dataset is not available for download.

\subsection{Hockey}

The Hockey Fight dataset~\cite{bermejo2011violence} contains 1,000 video clips from National Hockey League (NHL) games for binary classification of fight and non-fight. The Player Tracklet dataset~\cite{vats2021player} consists of 84 video clips from NHL games with annotated bounding boxes and identity labels for players and referees and is suitable for player tracking and identification.

\subsection{Tennis}
Various datasets have been constructed for tennis video analysis. ACASVA~\cite{de2011evaluation} is designed for tennis action recognition and consists of six broadcast videos of tennis games with labeled player positions and time boundaries of actions. THETIS~\cite{gourgari2013thetis} includes 1,980 self-recorded videos of 12 tennis actions with RGB, depth, 2D skeleton, and 3D skeleton videos, which can be used for multiple types of action recognition models. TenniSet~\cite{faulkner2017tenniset} contains five Olympic tennis match videos with six labeled event categories and textural descriptions, making it suitable for both recognition, localization, and action retrieval tasks. 

It should be noted that some recent works focus more on generative tasks, like PVG~\cite{menapace2021playable}, which obtained a tennis dataset through YouTube videos. PE-Tennis~\cite{menapace2022playable} built upon PVG and introduces camera calibration resulting from reconstruction, making it possible to edit the viewpoint. LGEs-Tennis~\cite{menapace2023plotting} enables generation from text editing on player movement, shot type, and location.

\subsection{Table Tennis}
Various datasets have been developed for table tennis stroke recognition, such as TTStroke-21~\cite{martin2018sport}, which comprises 129 self-recorded videos of 21 categories, and SPIN~\cite{schwarcz2019spin}, which includes 53 hours of self-recorded videos with annotations of ball position and player joints. OpenTTGames~\cite{voeikov2020ttnet} consists of 12 HD videos of table tennis games, labeled with ball coordinates and events. Stroke Recognition~\cite{kulkarni2021table} is similar to TTStroke-21, but much larger, and P$^2$A~\cite{bian2022textbf} is one of the largest datasets for table tennis analysis, with annotations of each stroke in 2,721 broadcasting videos.

\subsection{Gymnastics}
The FineGym~\cite{shao2020finegym} is a recent work developed for gymnastic action recognition and localization. It contains 303 videos with around 708-hour length and is annotated hierarchically, making it suitable for fine-grained action recognition and localization. On the other hand, AFG-Olympics~\cite{zahan2023learning} provides challenging scenarios with extensive background, viewpoint, and scale variations over an extended sample duration of up to 2 minutes. Additionally, a discriminative attention module is proposed to embed long-range spatial and temporal correlation semantics.

\subsection{Badminton}
The Badminton Olympic~\cite{ghosh2018towards} provides annotations for player detection, point localization, action recognition, and localization tasks. It comprises 10 YouTube videos of singles badminton matches, each approximately an hour long. The dataset includes annotations for player positions, temporal locations of point wins, and time boundaries and labels of strokes. Meanwhile, Stroke Forecasting~\cite{wang2021shuttlenet} contains 43,191 trimmed video clips of badminton strokes categorized into 10 types, which can be used for both action recognition and stroke forecasting.

\subsection{Figure skating}

There are 5 datasets proposed for figure skating action recognition in recent years. 
FineSkating~\cite{shan2020fineskating} is a hierarchical-labeled dataset of 46 videos of figure skating competitions for action recognition and action quality assessment. FSD-10~\cite{liu2020fsd} comprises ten categories of figure skating actions and provides scores for action quality assessment. FisV-5~\cite{FisV-5} is a dataset of 500 figure skating competition videos labeled with scores by 9 professional judges. FR-FS~\cite{TSA-Net} is designed to recognize figure skating falls, with 417 videos containing the movements of take-off, rotation, and landing. MCFS~\cite{liu2021temporal} has three-level annotations of figure skating actions and their time boundaries, allowing for action recognition and localization.

\subsection{Diving}
There are three diving datasets available for action recognition and action quality assessment. Diving48~\cite{li2018resound} contains 18,404 video segments covering 48 fine-grained categories of diving actions, making it a relatively low-bias dataset suitable for model evaluation. In contrast, MTL-AQA~\cite{parmar2019and} consists of 1,412 samples annotated with action quality scores, class labels, and textural commentary, making it suitable for multiple tasks. In addition, FineDiving~\cite{xu2022finediving} is a recent dataset consisting of 3,000 video samples covering 52 types of actions, 29 sub-action types, and 23 difficulty levels, providing fine-grained annotations including action types, sub-action types, coarse and fine time boundaries, and action scores. It is the first fine-grained motion video dataset for the AQA task, filling the gap in fine-grained annotations in AQA and suitable for designing competition strategies and better showcasing athletes' strengths.

\subsection{Dance}
The field of deep learning has several research tasks for dance, including music-oriented choreography, dance motion synthesis, and multiple object tracking. Researchers propose several datasets to promote research in this field. GrooveNet~\cite{alemi2017groovenet} consists of approximately 23 minutes of motion capture data recorded at 60 frames per second and four performances by a dancer. Dance with Melody~\cite{tang2018dance} includes 40 complete dance choreographies for four types of dance, totaling 907,200 frames collected with optical motion capture equipment. EA-MUD~\cite{sun2020deepdance} includes 104 video sequences of 12 dancing genres, while AIST++~\cite{li2021learn} is a large-scale 3D human dance motion dataset with frame-level annotations including 9 views of camera intrinsic and extrinsic parameters, 17 COCO-format human joint locations in both 2D and 3D, and 24 SMPL pose parameters. These datasets can be used for tasks such as dance motion recognition, tracking, and quality assessment.

\subsection{Sport Related Datasets for General Purpose} 
There are several datasets for sports action recognition and assessment tasks, including UCF sports~\cite{rodriguez2008action}, MSR Action3D~\cite{li2010action}, Olympic~\cite{niebles2010modeling}, Sports 1M\cite{karpathy2014large}, SVW~\cite{safdarnejad2015sports}, MultiSports~\cite{li2021multisports}, OlympicSports~\cite{pirsiavash2014assessing}, OlympicScoring~\cite{parmar2017learning}, and AQA~\cite{parmar2019action}. These datasets cover different sports, including team sports and individual sports, and provide various annotations, such as action labels, quality scores, and bounding boxes. 

Additionally, Win-Fail~\cite{parmar2022win} is a dataset specifically designed for recognizing the outcome of actions, while SportsPose~\cite{ingwersen2023sportspose} is the largest markerless dataset for 3D human pose estimation in sports, containing 5 short sports-related activities recorded from 7 cameras, totaling 1.5 million frames. SportsMOT~\cite{cui2023sportsmot} is a large-scale and high-quality multi-object tracking dataset comprising detailed annotations for each player present on the field in diverse sports scenarios. These datasets provide valuable resources for researchers to develop and evaluate algorithms for various sports-related tasks.

\subsection{Others}
CVBASE Handball~\cite{pers2005cvbase} and CVBASE Squash~\cite{pers2005cvbase} are developed for handball and squash action recognition, respectively, with annotated trajectories of players and action categories. GolfDB~\cite{mcnally2019golfdb} facilitates the analysis of golf swings, providing 1,400 high-quality golf swing video segments, action labels, and bounding boxes of players. Lastly, FenceNet~\cite{zhu2022fencenet} consists of 652 videos of expert-level fencers performing six categories of actions, with RGB frames, 3D skeleton data, and depth data provided. Rugby sevens~\cite{maglo2022efficient} is a public sports tracking dataset with tracking ground truth and the generated tracks. MLB-YouTube\cite{piergiovanni2018fine} is introduced for fine-grained action recognition in baseball videos.

\section{Virtual Environments}\label{sec:env}

Researchers can utilize virtual environments for simulation. In a virtual environment that provides agents with simulated motion tasks, multiple data information can be continuously generated and retained in the simulation. For example, Fever Basketball~\cite{jia2020fever} is an asynchronous environment, which supports multiple characters, multiple positions, and both the single-agent and multi-agent player control modes.

There are many virtual soccer games, such as rSoccer~\cite{martins2021rsoccer}, RoboCup Soccer Simulator~\cite{kitano1997robocup}, the DeepMind MuJoCo Multi-Agent Soccer Environment~\cite{liu2019emergent,liu2021motor} and JiDi Olympics Football~\cite{jidi_football}. rSoccer~\cite{martins2021rsoccer} and JiDi Olympics Football~\cite{jidi_football} are two toy football games in which plays are just rigid bodies and can just move and push the ball. However, players in GFootball~\cite{kurach2020google} have more complex actions, such as dribbling, sliding, and sprinting. Besides, environments like RoboCup Soccer Simulator~\cite{kitano1997robocup} and DeepMind MuJoCo Multi-Agent Soccer Environment~\cite{liu2019emergent,liu2021motor} focus more on low-level control of a physics simulation of robots, while GFootball focuses more on developing high-level tactics. 
To improve the flexibility and control over environment dynamics, SCENIC~\cite{azad2022programmatic} is proposed to model and generate diverse scenarios in a real-time strategy environment programmatically.


\section{Challenges}\label{sec:challenges}
In recent years, deep learning has emerged as a powerful tool in the analysis and enhancement of sports performance. The application of these advanced techniques has revolutionized the way athletes, coaches, and teams approach training, strategy, and decision-making. By leveraging the vast amounts of data generated in sports, deep learning models have the potential to uncover hidden patterns, optimize performance, and provide valuable insights that can inform decision-making processes. However, despite its promising potential, the implementation of deep learning in sports performance faces several challenges that need to be addressed to fully realize its benefits.

\paragraph{\textbf{Task Challenge}}
The complex and dynamic nature of sports activities presents unique challenges for computer vision tasks in tracking and recognizing athletes and their movements. Issues such as identity mismatch due to similar appearances~\cite{peize2021dance, zhang2020multi}, blurring~\cite{8909871} caused by rapid motion, and occlusion~\cite{naik2022deepplayer, naik2023yolov3} from other players or objects in the scene can lead to inaccuracies and inconsistencies in tracking and analysis. 
Developing robust and adaptable algorithms that can effectively handle these challenges is essential to improve the performance and reliability of deep learning models in sports applications.

\paragraph{\textbf{Datasets Standardization}}
Standardizing datasets for various sports is a daunting task, as each sport has unique technical aspects and rules that make it difficult to create a unified benchmark for specific tasks. For example, taking action recognition tasks as an example, in diving~\cite{xu2022finediving}, only the movement of the athlete needs to be focused on, and attention should be paid to the details of role actions. However, in team sports such as volleyball~\cite{ibrahim2016hierarchical}, more attention is needed to distinguish and identify targets and cluster the same actions after identification. Given the varying emphases of tasks, there are substantial differences in the dataset requirements. To go further, action recognition of the same sport type, involves nuanced differences in label classification, making it challenging to develop a one-size-fits-all solution or benchmark. 
The creation of standardized, user-friendly, open-source, high-quality, and large-scale datasets is crucial for advancing research and enabling fair comparisons between different models and approaches in sports performance analysis.

\paragraph{\textbf{Data Utilization}}
The sports domain generates vast amounts of fine-grained data through sensors and IoT devices. However, current data processing methods primarily focus on computer vision and do not fully exploit the potential of end-to-end deep learning approaches. To fully harness the power of these rich data sources, researchers must develop methods that combine fine-grained sensor data with visual information. This fusion of diverse data streams can enable more comprehensive and insightful analysis, leading to significant advancements in the field of sports performance. Some studies have shown that introducing multi-modal data can benefit the analysis of athletic performance. For example, in table tennis, visual and IOT signals can be simultaneously used to analyze athlete performance~\cite{wang2022tac}. In dance, visual and audio signals are both important~\cite{li2021learn}. 
More attention is needed on how to utilize diverse data, so as to achieve better fusion. Meanwhile, multi-modal algorithms and datasets~\cite{li2021learn} are both necessary.

\section{Future trend}\label{sec:fw}
The integration of deep learning methodologies into sports analytics can empower athletes, coaches, and teams with unprecedented insights into performance, decision-making, and injury prevention. This future work aims to explore the transformative impact of deep learning techniques in sports performance, focusing on data generation methods, multi-modality and multi-task models, foundation models, applications, and practicability.

\paragraph{\textbf{Multi-modality and Multi-task}}
By harnessing the power of multi-modal data and multi-task learning, robust and versatile models capable of handling diverse and complex sports-related challenges can be fulfilled. Furthermore, we will investigate the potential of large-scale models in enhancing predictive and analytical capabilities. It consists of practical applications and real-world implementations that can improve athlete performance and overall team dynamics. Ultimately, this work seeks to contribute to the growing body of research on deep learning in sports performance, paving the way for novel strategies and technologies that can revolutionize the world of sports analytics.

\paragraph{\textbf{Foundation Model}}    
The popularity of ChatGPT has demonstrated the power of large language models~\cite{liu2023summary}, while the recent segment-anything project showcases the impressive performance of large models in visual tasks~\cite{kirillov2023segment}. The prompt-based paradigm is highly capable and flexible in natural language processing and even image segmentation, offering unprecedented rich functionality. For example, some recent work has leveraged segment-anything in medical image~\cite{deng2023segment,roy2023sammd,liu2023samm}, achieving promising results by providing point or bounding box prompts for preliminary zero-shot capability assessment, demonstrating that segment anything model~(SAM) has good generalization performance in medical imaging. Therefore, the development of large models in the sports domain should consider how to combine existing large models to explore applications, and how to create large models specifically for the sports domain.

Combining large models requires considering the adaptability of the task. Compared to the medical field, sports involve a high level of human participation, inherently accommodating different levels and modalities of methods and data. We believe that both large language models in natural language processing and large image segmentation models in computer vision should have strong compatibility in sports. In short, we believe there is potential for exploring downstream tasks, such as using ChatGPT for performance evaluation and feedback: employ ChatGPT to generate natural language summaries of player or team performance, as well as provide personalized feedback and recommendations for improvement.

Foundation models directly related to the sports domain require a vast amount of data corresponding to the specific tasks. For visual tasks, for example, it is essential to ensure good scalability, adopt a prompt-based paradigm, and maintain powerful capabilities while being flexible and offering richer functionality. It is important to note that large models do not necessarily imply a large number of parameters, but rather a strong ability to solve tasks. Recent work on segment-anything has proven that even relatively simple models can achieve excellent performance when the data volume is sufficiently large. Therefore, creating large-scale, high-quality datasets in the sports domain remains a crucial task.

\paragraph{\textbf{Data Generation}}
High-quality generated data can significantly reduce manual labor costs while demonstrating the diversity that generative models can bring. Many studies~\cite{li2021learn, wu2022tune} have focused on generating sports videos, offering easily editable, high-quality generation methods, which are elaborated upon in the relevant Section~\ref{sec:PMG} and \ref{sec:MG}. 
Meanwhile, by combining large models, additional annotation work can be performed at this stage, and if possible, new usable data can be generated.

\paragraph{\textbf{Applications}}
Though there are many excellent automatic algorithms for different tasks in the field of sports, they are still insufficient when it comes to deployment for specific tasks. In the daily exercise of ordinary people, who generally lack professional guidance, there should be more applications that make good use of these deep learning algorithms, and use more user-friendly and intelligent methods to promote sports for everyone. There are already some works~\cite{liu2022posecoach, zhao2020human, perrett2022personalized} focusing on sports performance analysis, data recording visualization, energy expenditure estimation, and many other aspects. At the same time, in professional sports, there are also some works~\cite{wang2022tac, wang2019ai} that focus on combining various data and methods to help improve athletic performance. Broadly speaking, in both daily life and professional fields, there is a need for more applications relating to health and fitness assessments.

\paragraph{\textbf{Practicability}}
In more challenging, high-level tasks with real-world applications, practicality becomes increasingly important. Many practical challenges remain unexplored or under-explored in applying deep learning to sports performance. In decision-making, for example, current solutions often rely on simulation-based approaches. However, multi-agent decision-making techniques hold great potential for enhancing real-world sports decision-making. Tasks such as ad-hoc teamwork~\cite{radke2023presenting} in multi-agent systems and zero-shot human-machine interaction are crucial for enabling effective and practical real-world applications. Further research is needed to bridge the gap between theoretical advancements and their practical implications in sports performance analysis and decision-making. For example, RoboCup~\cite{kitano1997robocup} aims to defeat human players in the World Cup by 2050. This complex task requires robots to perceive their environment, gather information, understand it, and execute specific actions. Such agents must exhibit sufficient generalization, engage in extensive human-machine interaction, and quickly respond to performance and environmental changes in real-time.

\section{Conclusion}\label{sec:conc}
In this paper, we present a comprehensive survey of deep learning in sports, focusing on four main aspects: algorithms, datasets, challenges, and future works. We innovatively summarize the taxonomy and divide methods into perception, comprehension, and decision from low-level to high-level tasks. In the challenges and future works, we provide cutting-edge methods and give insights into the future trends and challenges of deep learning in sports.

\section*{Acknowledgments}
This work is supported by National Key R\&D Program of China under Grant No.2022ZD0162000, and National Natural Science Foundation of China No.62106219.


\bibliographystyle{IEEEtran}

\bibliography{ref}
\begin{IEEEbiography}[{\includegraphics[height=1.25in,clip,keepaspectratio]{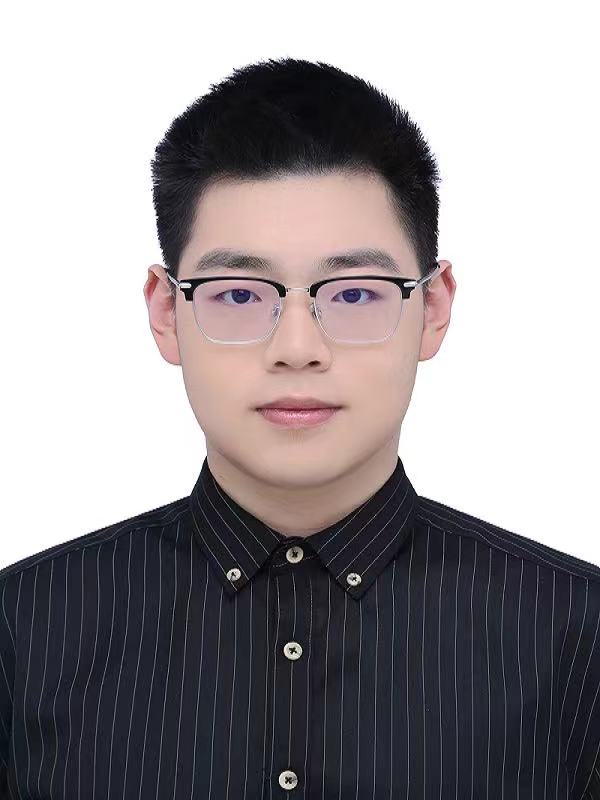}}]{Zhonghan Zhao}
received the BE degree from Communication University of China. He is currently working toward the PhD degree with Zhejiang University - University of Illinois Urbana-Champaign Institute, Zhejiang University. His research interests include machine learning, reinforcement learning and computer vision.
\end{IEEEbiography}

\begin{IEEEbiography}[{\includegraphics[height=1.25in,clip,keepaspectratio]{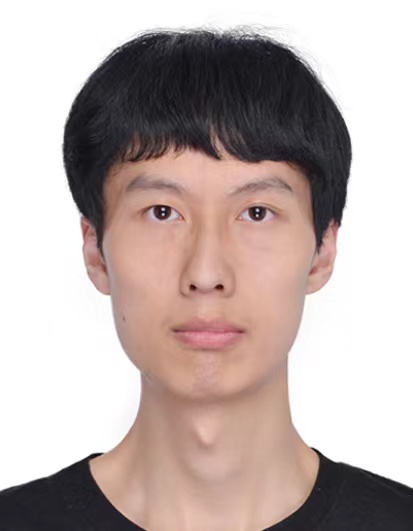}}]{Wenhao Chai}
received the BE degree from Zhejiang University, China. He is currently working toward the Master degree with University of Washington. His research interests include 3D human pose estimation, generative models, and multi-modality learning.
\end{IEEEbiography}

\begin{IEEEbiography}[{\includegraphics[height=1.25in,clip,keepaspectratio]{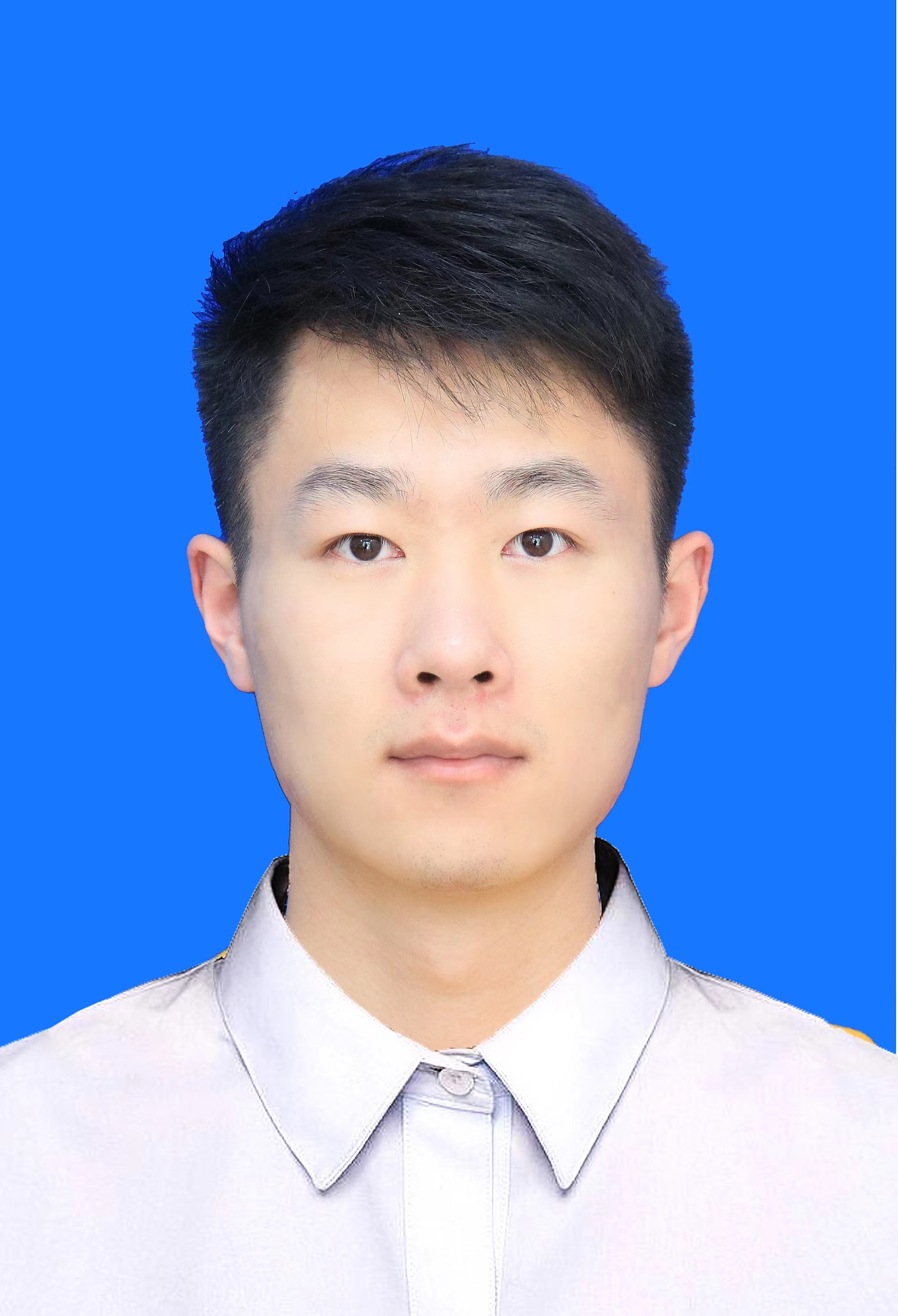}}]{Shengyu Hao}
received the MS degree from Beijing University of Posts and Telecommunications, China. He is currently working toward the PhD degree with Zhejiang University - University of Illinois Urbana-Champaign Institute, Zhejiang University. His research interests include machine learning and computer vision.
\end{IEEEbiography}

\begin{IEEEbiography}[{\includegraphics[height=1.25in,clip,keepaspectratio]{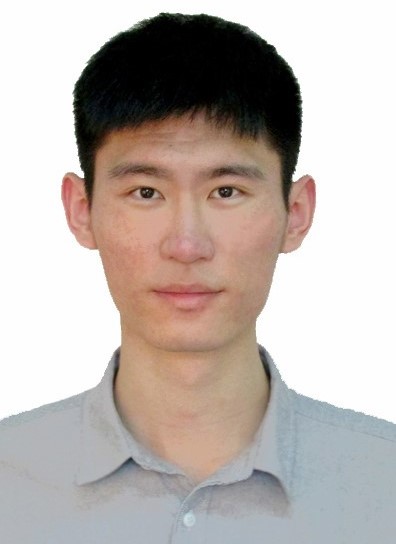}}]{Wenhao Hu}
received the BS degree from Zhejiang University, China. He is currently working toward the PhD degree with Zhejiang University - University of Illinois Urbana-Champaign Institute, Zhejiang University. His research interests include generative models and 3D reconstruction.
\end{IEEEbiography}

\begin{IEEEbiography}[{\includegraphics[height=1.25in,clip,keepaspectratio]{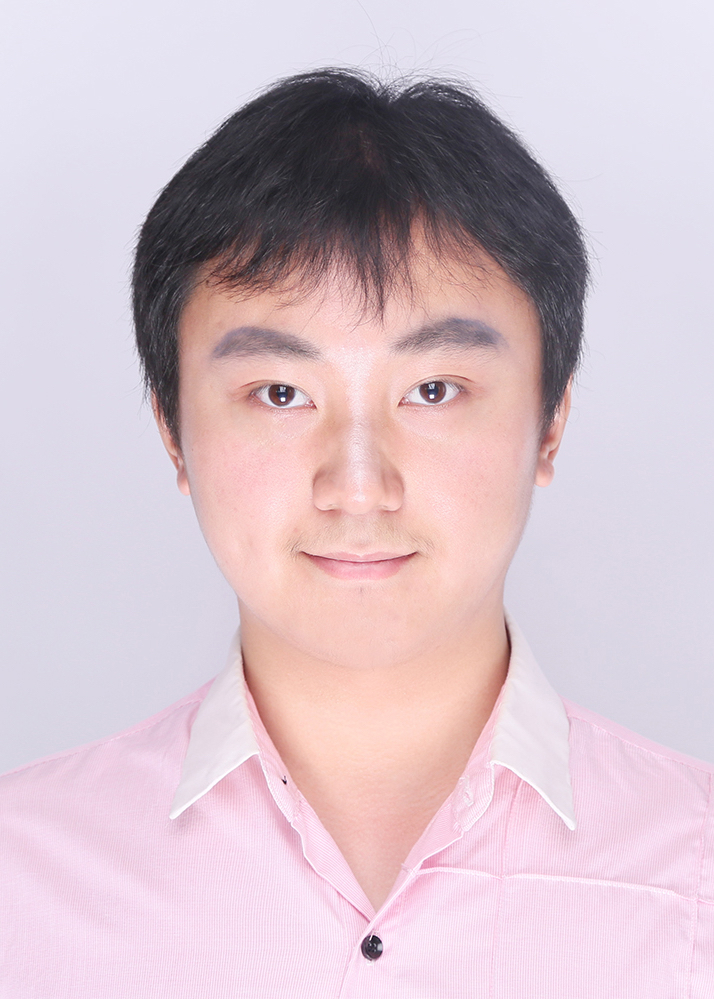}}]{Guanhong Wang}
received the MS degree from Huaqiao University, China. He is currently working toward the PhD degree with Zhejiang University - University of Illinois Urbana-Champaign Institute, Zhejiang University. His research interests include deep learning and computer vision.
\end{IEEEbiography}

\begin{IEEEbiography}[{\includegraphics[height=1.25in,clip,keepaspectratio]{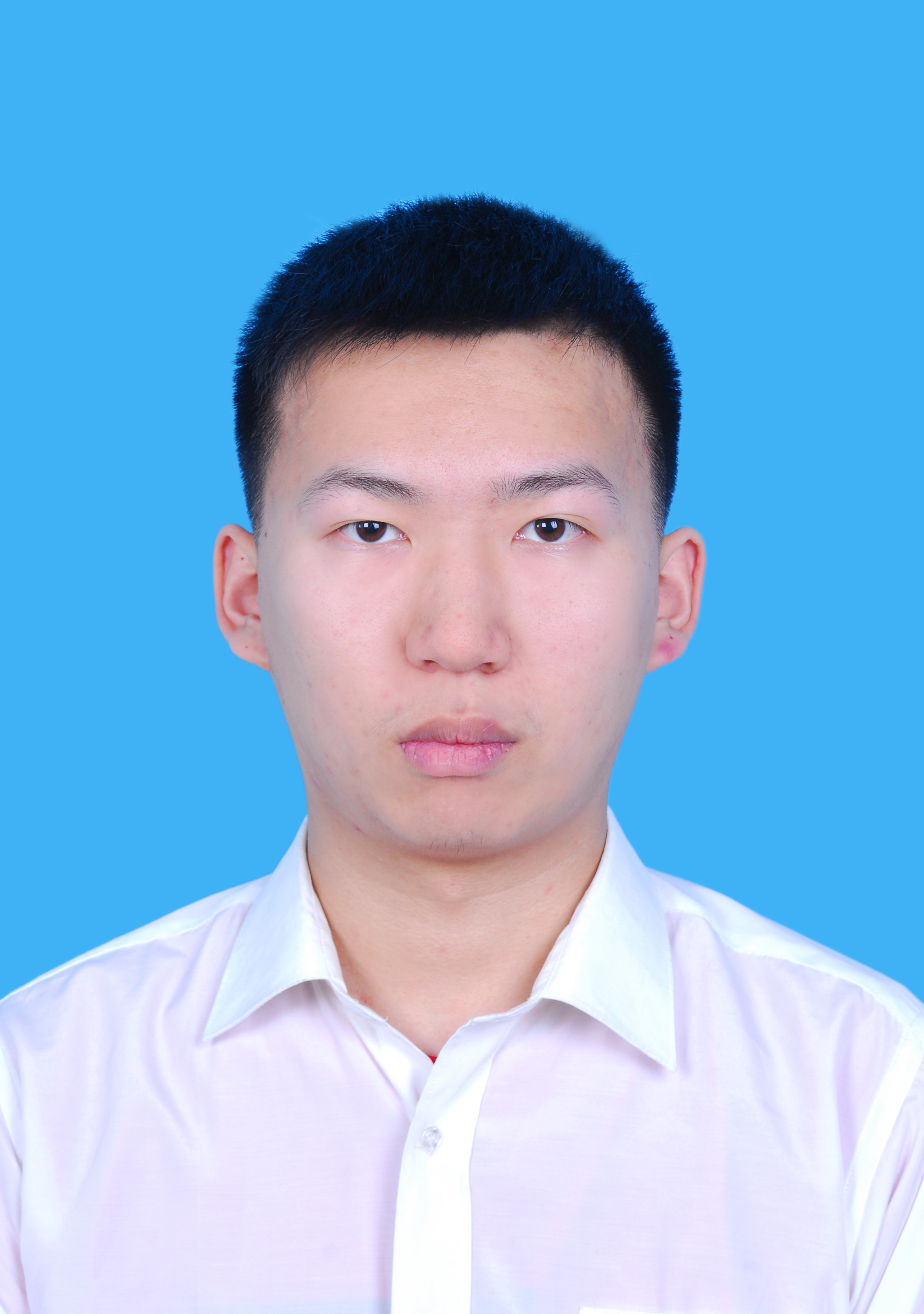}}]{Shidong Cao}
received the BE degree from Beijing University of Posts and Telecommunications, China. He is currently working toward the MS degree with Zhejiang University - University of Illinois Urbana-Champaign Institute, Zhejiang University. His research interests include machine learning and computer vision.
\end{IEEEbiography}

\begin{IEEEbiography}[{\includegraphics[width=1in,height=1.25in,keepaspectratio]{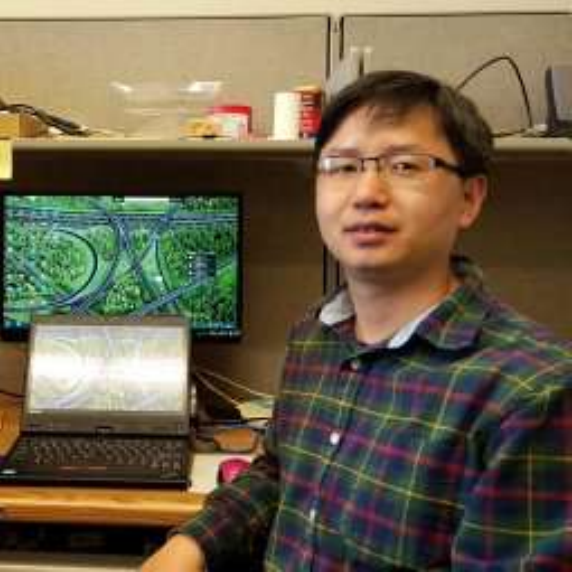}}]{Dr. Mingli Song}
received the Ph.D. degree in computer science from Zhejiang University, China, in 2006. He is currently a Professor with the Microsoft Visual Perception Laboratory, Zhejiang University. His research interests include face modeling and facial expression analysis. He received the Microsoft Research Fellowship in 2004.
\end{IEEEbiography}

\begin{IEEEbiography}[{\includegraphics[width=1in,height=1.25in,keepaspectratio]{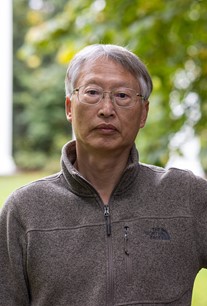}}]{Dr. Jenq-Neng Hwang}
received the BS and MS degrees, both in electrical engineering from the National Taiwan University, Taipei, Taiwan, in 1981 and 1983 separately. He then received his Ph.D. degree from the University of Southern California. In the summer of 1989, Dr. Hwang joined the Department of Electrical and Computer Engineering (ECE) of the University of Washington in Seattle, where he has been promoted to Full Professor since 1999. He is the Director of the Information Processing Lab. (IPL), which has won several AI City Challenges and BMTT Tracking awards in the past years. Dr. Hwang served as associate editors for IEEE T-SP, T-NN and T-CSVT, T-IP and Signal Processing Magazine (SPM). He was the General Co-Chair of 2021 IEEE World AI IoT Congress, as well as the program Co-Chairs of IEEE ICME 2016, ICASSP 1998 and ISCAS 2009. Dr. Hwang is a fellow of IEEE since 2001.
\end{IEEEbiography}

\begin{IEEEbiography}[{\includegraphics[width=1in,height=1.25in,clip,keepaspectratio]{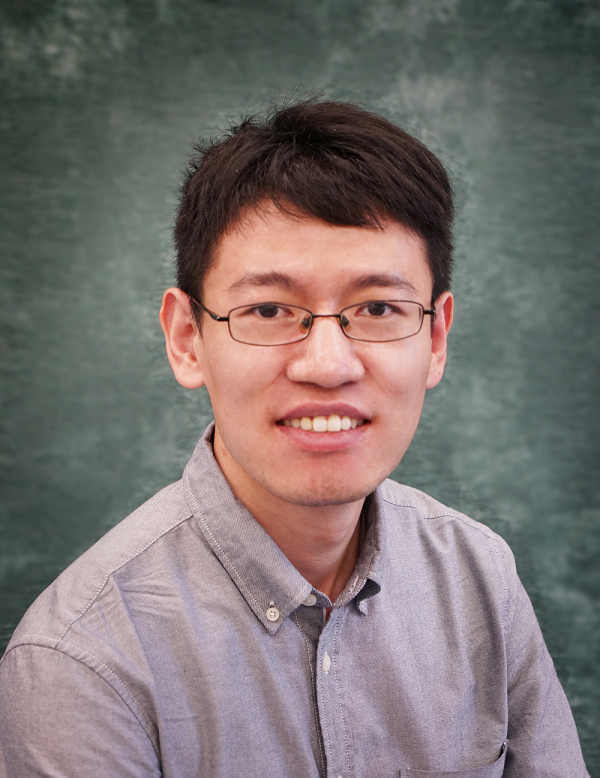}}]{Dr. Gaoang Wang}
joined the international campus of Zhejiang University as an Assistant Professor in September 2020. He is also an Adjunct Assistant Professor at UIUC. Gaoang Wang received a B.S. degree at Fudan University in 2013, a M.S. degree at the University of Wisconsin-Madison in 2015, and a Ph.D. degree from the Information Processing Laboratory of the Electrical and Computer Engineering department at the University of Washington in 2019. After that, he joined Megvii US office in July 2019 as a research scientist working on multi-frame fusion. He then joined Wyze Labs in November 2019 working on deep neural network design for edge-cloud collaboration. His research interests are computer vision, machine learning, artificial intelligence, including multi-object tracking, representation learning, and active learning. Gaoang Wang published papers in many renowned journals and conferences, including IEEE T-IP, IEEE T-MM, IEEE T-CSVT, IEEE T-VT, CVPR, ICCV, ECCV, ACM MM, IJCAI.
\end{IEEEbiography}

\end{document}